\pdfoutput=1

\documentclass[11pt, dvipsnames, table]{article}

\usepackage[]{acl}

\usepackage{times}
\usepackage{latexsym}

\usepackage[T1]{fontenc}

\usepackage[utf8]{inputenc}

\usepackage{microtype}
\usepackage{graphicx}
\usepackage{amsmath}
\usepackage{bbm}
\usepackage{amssymb}
\usepackage{amsfonts}
\usepackage{booktabs}
\usepackage{lipsum}
\usepackage{multicol}
\usepackage{makecell}
\usepackage{rotating}
\usepackage{multirow}
\usepackage{cleveref}
\usepackage{inconsolata}
\usepackage{xcolor}
\usepackage[stable]{footmisc}

\newcommand{\tfive}{\textsc{T5}}
\newcommand{\tl}{\textsc{T5-large}}
\newcommand{\txxl}{\textsc{T5-xxl}}
\newcommand{\ktfive}{\textsc{KniT5}}
\newcommand{\ktl}{\textsc{KniT5-large}}
\newcommand{\ktxxl}{\textsc{KniT5-xxl}}
\newcommand{\hp}{Hopping Prompts}
\newcommand{\modular}{Parse-then-Hop}
\newcommand{\spot}{\textsc{SPoT}}
\newcommand{\pth}{\textsc{PaTH}}
\newcommand{\mixture}{\textsc{MixHop}}
\newcommand{\nwalks}{165{,}324}
\newcommand{\wpn}{20}
\newcommand{\walktrain}{155{,}311}
\newcommand{\ntrain}{72{,}759}
\newcommand{\ndev}{8{,}085}
\newcommand{\ntest}{6{,}768}
\newcommand{\walkdev}{8{,}085}
\newcommand{\walktest}{6{,}768}
\newcommand{\ents}{98{,}284}
\newcommand{\onehopqa}{1WikiHopQA}

\newcommand{\correctans}[1]{\textcolor{OliveGreen}{\textbf{#1}}}
\newcommand{\wrongans}[1]{\textcolor{BrickRed}{\textbf{#1}}}

\newcommand\blankfootnote[1]{%
  \let\thefootnote\relax\footnotetext{#1}%
  \let\thefootnote\svthefootnote%
}

\title{Triggering Multi-Hop Reasoning for Question Answering\\in Language Models using Soft Prompts and Random Walks}

\author{Kanishka Misra$^\bigstar$\\
  Purdue University \\
  \texttt{kmisra@purdue.edu}\\
  \And
  Cicero Nogueira dos Santos\\
  Google Research \\
  \texttt{cicerons@google.com}\\
  \And
  Siamak Shakeri\\
  Google DeepMind \\
  \texttt{siamaks@google.com}
 }

\begin{document}
\maketitle
\begin{abstract}
{Despite readily memorizing world knowledge about entities, pre-trained language models (LMs) struggle to compose together two or more facts to perform multi-hop reasoning in question-answering tasks.
In this work, we propose techniques that improve upon this limitation by relying on random walks over structured knowledge graphs. 
Specifically, we use soft prompts to guide LMs to chain together their encoded knowledge by learning to map multi-hop questions to random walk paths that lead to the answer. 
Applying our methods on two T5 LMs shows substantial improvements over standard tuning approaches in answering questions that require 2-hop reasoning.}
\blankfootnote{$\bigstar$ Work done during an internship at Google Research.}
\end{abstract}

\section{Introduction}
Performing multi-hop reasoning to answer questions such as \textit{Where was David Beckham’s daughter born?} requires two fundamental capacities:
\textbf{C1:} possessing pre-requisite knowledge (\textit{David Beckham’s daughter is Harper Beckham}, \textit{Harper Beckham was born in Los Angeles}), and \textbf{C2:} ability to compose internalized knowledge.
Contemporary pre-trained language models (LMs) such as BERT \citep{devlin-etal-2019-bert} and T5 \citep{raffel2020exploring} have been shown to be adept at encoding factual knowledge \citep{petroni-etal-2019-language, zhong-etal-2021-factual, roberts-etal-2020-much}, an ability that can be further boosted by explicitly integrating them with knowledge about entities and relations \citep[\textit{i.a.}]{bosselut-etal-2019-comet, sun-etal-2020-colake, wang-etal-2021-kepler}.
At the same time, these LMs often struggle to compose the knowledge they encode \citep{kassner-etal-2020-pretrained, talmor-etal-2020-olmpics, moiseev-etal-2022-skill}, and therefore do not satisfy \textbf{C2}.
To overcome this limitation, previous works have proposed methods that decompose multi-hop questions into single hop sub-questions that models can more easily answer \citep[\textit{i.a.}]{min-etal-2019-multi, perez-etal-2020-unsupervised}. However, such methods require training entirely separate models, or make use of human-annotations \citep{patel2022question}. Furthermore, they focus on tasks where models explicitly receive additional text containing relevant facts, which makes it unclear if they can \textit{truly} compose the knowledge that they have internalized.

In this work, we aim to improve the standalone, self-contained ability of LMs to perform multi-hop reasoning. We posit that \emph{random walks}---paths between entity nodes sampled from structured knowledge graphs---can provide a useful training signal for LMs to compose entity knowledge. To test this, we perform a case-study on two T5 models \citep[\textsc{large} and \textsc{xxl,}][]{raffel2020exploring}. 
Specifically, we first integrate within the LMs the single-hop knowledge that is required to answer multi-hop questions (effectively guaranteeing \textbf{C1} is met). 
We show that this alone is not enough to demonstrate substantial improvements on questions requiring 2-hop reasoning.
{We then adapt the knowledge integrated T5 models by training soft prompts \citep{qin-eisner-2021-learning, lester-etal-2021-power} on random walks over the structured knowledge that they have encoded, and devise two methods that trigger this ability in the LMs given a multi-hop question as input. 
The first method, \textbf{\modular{}} (\pth), uses two specialized soft prompts: one to parse entities and relations from the question, and another to generate a path to the answer, resembling the outputs of a random walk. The second method, \textbf{\mixture}, trains a single prompt on a mixture that combines the QA task with the random walk training, so as to allow the model to implicitly learn \pth's task. Both these soft prompt methods use the same underlying LM (kept frozen), and guide it to compose its internalized entity knowledge.}

\begin{figure*}[t!]
    \centering
    \includegraphics[width=\textwidth]{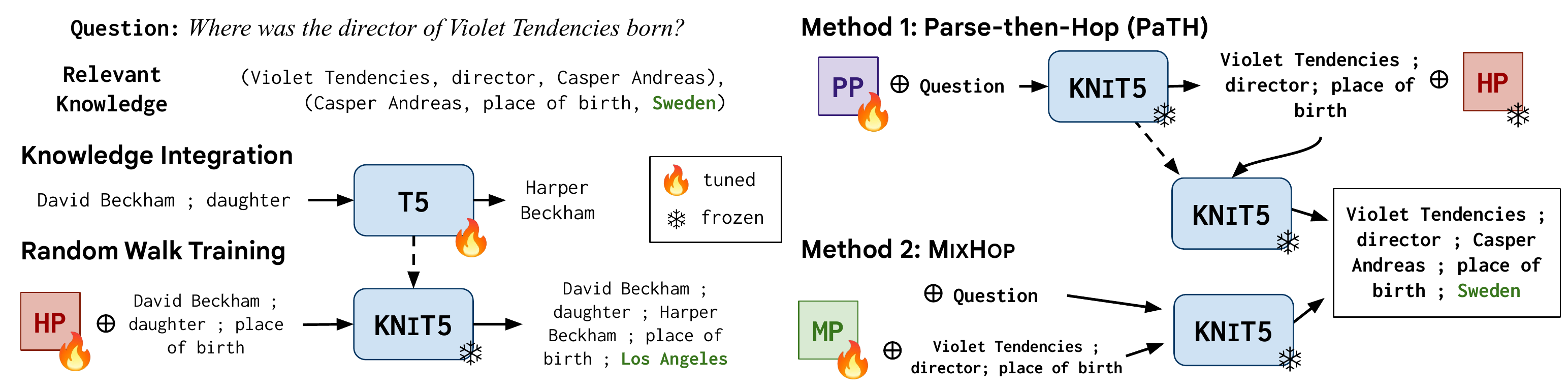}
    \caption{Overview of our approach. Colored rectangular boxes indicate soft prompts: Hopping Prompts (HP), Parsing Prompts (PP), and Prompts for the \mixture{} approach (MP). $\bigoplus$ indicates concatenation.}
    \label{fig:method}
    \vspace{-1em}
\end{figure*}

Our experiments suggest that integrating random walks in the \tfive{} models using our proposed techniques can substantially improve their ability to answer entity-centric 2-hop questions \citep{ho-etal-2020-constructing} at larger model sizes.
Briefly, on \txxl{} our methods show improvements over previously proposed prompt-tuning approaches \citep{lester-etal-2021-power, vu-etal-2022-spot} as well as full model fine-tuning, with \pth{} and \mixture{} demonstrating gains of $\sim$16 and $\sim$9.6 points in exact match scores over fine-tuning the entire model, respectively.
In the case of \tl{}, our methods demonstrate improvements over standard prompt-tuning methods, but fall short of the performance achieved using fine-tuning, suggesting that larger models---with up to 11B parameters---are more conducive to leveraging the training signal provided by random walks via soft prompts.

\section{Method}

\subsection{Models} We apply our methods on two T5.1.1 models 
\citep{raffel2020exploring}---\tl{} (770M parameters) and \txxl{} (11B parameters), using checkpoints that have been adapted using the Prefix LM objective for 100K steps \citep{lester-etal-2021-power}.

\subsection{Knowledge Integration}
\label{sec:ki}
We first ensure that the LMs we use have the prerequisite single-hop knowledge (C1) required to answer multi-hop questions. This is necessary, as preliminary experiments suggested that the T5 models we used did not satisfy this primary criterion for multi-hop reasoning (see \Cref{tab:onehopresults}). 
Specifically, we follow \citet{bosselut-etal-2019-comet} and fine-tune our LMs on knowledge graph (KG) triples containing the relevant knowledge that is to be composed to answer questions. 
That is, given a triple $(e_1, r, e_2)$, where $e_1$ and $e_2$ are entities, and $r$ is the relation, we fine-tune our T5 models to take as input the string ``\texttt{e1 ; r1}'', and produce ``\texttt{e2}'' as output, using the Prefix LM objective \citep{raffel2020exploring}. 
To avoid catastrophic forgetting \citep{mccloskey1989catastrophic} and retain the LMs' language understanding abilities, we mix our knowledge integration training instances with that of the models' pre-training corpus---i.e., C4 \citep{raffel2020exploring}---in a 50:50 mixture.
We denote the resulting models as \textbf{KN}owledge-\textbf{I}ntegrated \textbf{T5} (\ktfive{}).

\subsection{Composing knowledge using soft prompts}
\paragraph{Random Walk training} 
Our method is centered around guiding the \ktfive{} LMs to chain together their encoded knowledge by training them on random walks over a relevant KG. We formulate random walks here as as a sequence of entity-relation-entity triples that are connected linearly via shared entities. Figure \ref{fig:method} shows an example with a random walk of length 3 (\texttt{Violet Tendencies ; director ; Casper Andreas ; place of birth ; Sweden}). 
To perform our random walk training, we rely on soft prompts \citep{li-liang-2021-prefix, lester-etal-2021-power, qin-eisner-2021-learning}, a sequence of learnable token-vectors that are prepended to the input of the LM. 
Importantly, we only update these vectors during training, thereby keeping intact the utility and encoded knowledge of the main LM, while also being parameter efficient.
Our training procedure is as follows: we first perform uniform random walks of length $n$ over the KG used in \cref{sec:ki}, resulting in a set whose elements are sequences of entities interleaved by the relations that connect them: $(e_1, r_1, e_2, \dots, r_{n-1}, e_n)$.
During training, \ktfive{} receives as input an incomplete path, with only the initial entity and the intermediate relations $(e_1, r_1, r_2, \dots, r_{n-1})$, and is tasked to generate the full path: $(e_1, r_1, e_2, r_2 \dots, r_{n-1}, e_n)$.
We denote the trained prompts that trigger this ability in \ktfive{} as \textbf{\hp{}}.

\subsection{Performing QA using \hp}
\noindent
We propose two new techniques that utilize \hp{} to map natural language questions to appropriate paths in the knowledge graph:

\paragraph{\modular{} (\pth)} We take advantage of the modularity of soft prompts, and distribute the responsibility of parsing the relational structure from questions and random walk querying using separate specialized prompts, keeping the underlying model the same. We train ``parsing'' prompts that parse questions to incomplete random walk queries, resembling the inputs to the \hp{} described above.
For instance, the question ``\textit{Where was David Beckham's daughter born?}'' is parsed to ``\texttt{David Beckham ; daughter ; place of birth}''.
We then swap the parsing prompts with the hopping prompts, using the outputs from the parsing step as inputs and then run inference to get a path from the entity in the question to the answer: ``\texttt{David Beckham ; daughter ; Harper Beckham ; place of birth ; \textcolor{OliveGreen}{\textbf{Los Angeles}}}'', as shown in Figure \ref{fig:method}. 
{We posit that parsing of the appropriate relational structure from the question should be easy and self-contained, since it only involves using the surface form of the question as opposed to invoking any external knowledge, which is delegated to \hp{}.}

\paragraph{\mixture} We propose to jointly train a single set of prompts on a mixture of the QA task and the \hp{} task (50:50), thereby halving the number of forward passes from the previous method. 
Our primary motivation here is to provide diverse training signals that get models to map questions to the structured knowledge that explicitly connects the entity in the question to the answer entity. 
Like \pth{}, \mixture{} directly produces random walk paths as output, as shown in Figure \ref{fig:method}.

\section{Experimental Setup}

\subsection{Data}

\paragraph{Multi-hop QA Dataset} 
While traditional multi-hop QA datasets provide additional paragraphs \citep{yang-etal-2018-hotpotqa, trivedi-etal-2022-musique} for models to reason over, we operate under the more challenging closed-book QA setting \citep{roberts-etal-2020-much}, where such contexts are omitted. Specifically, we use the ``compositional” and ``inference” subsets of the \textbf{2WikiMultiHopQA} dataset \citep{ho-etal-2020-constructing}, which contains 2-hop English questions focusing on \ents{} entities and 29 relations, sourced from WikiData \citep{vrandevcic2014wikidata}. We select this dataset as it uniquely provides the \textit{precise} structured knowledge that is required to answer each question, in the form of entity-relation-entity triples.\footnote{Works such as \citet{balachandran-etal-2021-investigating} propose unsupervised mappings of questions in more popular datasets such as NaturalQuestions \citep{kwiatkowski-etal-2019-natural} to paths in knowledge graphs, but our initial investigations of these paths found them to be extensively noisy.}
Since the test splits for these specific subsets are private, we use the validation split as the test set, and use 10\% of the training set for validation. In total we have \ntrain{} train, \ndev{} validation, and \ntest{} test questions.

\paragraph{1-hop QA Dataset} To characterize if the models we test have the pre-requisite 1-hop knowledge, we additionally construct 1-hop questions from 2WikiMultiHopQA by applying manually defined templates over the entity triples provided for each 2-hop question (see Appendix \ref{sec:onehop}). 
For instance, the triple \texttt{Inception ; director ; Christopher Nolan} is converted to \textit{Who is the director of Inception?}. We end up with 83,643 train, 5,022 validation, and 6,440 test QA instances.
We term this constructed dataset as \textbf{\onehopqa}.

\paragraph{Knowledge Integration Data} 
We build the KG for our methods using the set of ground-truth triples provided in the 2WikiMultiHopQA dataset (\ents{} entities and 29 relations, amounting to 95K triples).

\paragraph{Random Walk Training Corpus} For each entity in the above KG, we sample \textit{up to} \wpn{} random walks of length 3, each corresponding to an instance of 2 hops between entities.
We repeat this step 5 times with different seeds, discard duplicate paths, and end up with a total of \nwalks{} unique paths as a result.
\textbf{Importantly, we hold out the paths that include the triples in the QA task's validation and test sets in order to avoid leakage}, ending up with \walktrain{}/ \walkdev{}/\walktest{} paths as our train/validation/test sets, respectively. 
This way, our experiments test for the kinds of generalization where models should successfully place entities in novel structures (complete paths in the KG), whose primitive knowledge (1-hop triples) is encoded in the model, but the composition is not. This can be viewed as a partial version of the lexical and structural generalization tests in stricter, more prominent compositional generalization benchmarks \citep{lake2018generalization, kim-linzen-2020-cogs}.

\subsection{Baselines and Comparisons} We compare our proposed approaches to standard fine-tuning and prompt-tuning \citep{lester-etal-2021-power}, which we use to directly produce the answer, without any intermediate entities or relations.
Additionally, we also adapt \spot{} \citep{vu-etal-2022-spot}, a prompt-tuning method where we initialize prompts with those that were pre-trained on related tasks. 
In our adaptation, we initialize prompts using the values of the \hp{}, and \spot-transfer them to guide \ktfive{} models to generate the full output, similar to \pth{} and \mixture{}. 
Since we operate in the closed book QA setting \citep{roberts-etal-2020-much}, our methods cannot be directly compared to previous approaches on the dataset we considered, all of which receive paragraph contexts during training.
Only two other methods have considered the present dataset in its closed-book format \citep{press2022measuring, wang2022iteratively}. However, both of them use smaller subsets of the validation set as their testing set, and test on different pre-trained models, making it impractical to directly compare our results to their reported values.

\section{Experiments and Findings\footnote{Training details for all experiments can be found in Appendix \ref{sec:trainingdeets}.}}

We report and summarize our results as follows:

\begin{table}[t!]
\centering
\begin{tabular}{@{}clcc@{}}
\toprule
\textbf{Setup} & \textbf{Model} & \textbf{\textsc{large}} & \textbf{\textsc{xxl}} \\ \midrule
\multirow{2}{*}{PT} & \tfive{} & 4.36 & 6.89 \\
 & \ktfive{} & \textbf{6.30} & \textbf{31.64} \\ \midrule
\multirow{2}{*}{FT} & \tfive{} & 6.24 & 8.82 \\
 & \ktfive{} & \textbf{22.73} & \textbf{43.60} \\ \bottomrule
\end{tabular}%
\caption{Test EM scores achieved by \tfive{} and \ktfive{} on \onehopqa{}. PT: Prompt-Tuning, FT: Fine-Tuning.}
\label{tab:onehopresults}
\end{table}

\paragraph{Integration of 1-hop knowledge only results in marginal improvements on 2-hop questions}
We begin by first establishing the extent to which T5 models encode and compose 1-hop knowledge required to answer 2-hop questions, and whether additional knowledge integration (via \ktfive) can improve both these abilities.
From Tables \ref{tab:onehopresults} and \ref{tab:twohopresults}, we observe that the \tfive{} models struggle to answer both 1-hop as well as 2-hop questions, suggesting that they critically lack the precise 1-hop entity knowledge required to demonstrate success on the 2-hop questions.
The \ktfive{} LMs overcome this limitation, by showing substantial gains on \onehopqa{} over their T5 counterparts---they show improvements of $\sim$16.5 and $\sim$34.8 points in exact match (EM) scores at \textsc{large} and \textsc{xxl} sizes in the fine-tuning setting, respectively (\Cref{tab:onehopresults}).
However, this is insufficient to show improvements on 2-hop questions---where maximum gain over \tfive{} is only 2.2 points, achieved by prompt-tuning \ktxxl{} (see \Cref{tab:twohopresults}).
This suggests that even after being endowed with the prerequisite 1-hop knowledge, both LMs are unable to successfully answer more complicated questions, echoing the results of \citet{moiseev-etal-2022-skill}.
Note that both \ktfive{} models almost perfectly memorize the KG in our knowledge-integration experiments (achieving $\sim$96\% EM in under 10K training steps; see \Cref{sec:aaki}), so their limitations on 2-hop questions are likely not due to lack of entity knowledge and perhaps instead due to the inability to compose or chain together memorized facts.

\paragraph{Generalizing to novel random walks may require the prompt-tuning of larger LMs}
We now turn to analyzing the performance of models in generating random walks, a critical component for all our proposed QA methods. How well does prompt-tuning LMs generalize to KG paths composed of facts they have memorized but are unseen during training?
Recall that this step involved leveraging soft prompts (called \hp{}) to guide the LMs to chain together their memorized entity knowledge and generate paths akin to performing a random walk. That is, it is the \hp{} that must provide the necessary condition in the encoder to facilitate successful output-generation, and not the entire LM. Also recall that we explicitly held out the paths involving triples in the validation and test sets of the main QA task to prevent complete memorization (due to leakage into the training set). 
This way we are able to measure the extent to which models learned to construct KG paths in a generalized manner. 
\renewcommand*{\arraystretch}{1.2}
\begin{table}[t!]
\centering
\begin{tabular}{@{}lcc@{}}
\toprule
\textbf{Model} & \textbf{EM} & \textbf{F1} \\ \midrule
\ktl{} & 22.83 & 84.72\\
\ktxxl{} & \textbf{58.36} & \textbf{92.82}\\ \bottomrule
\end{tabular}%
\caption{Best reported validation EM and F1 scores achieved from training \hp{} to get \ktfive{} models to generate random-walks. $N = 8085$.}
\label{tab:hpmetrics}
\end{table}
\renewcommand*{\arraystretch}{1.2}
\begin{table*}[t]
\centering
\begin{tabular}{@{}lccccccc@{}}
\toprule
\multirow{2}{*}{\textbf{Size}} & \multicolumn{2}{c}{\textbf{Prompt-Tuning}} & \multicolumn{2}{c}{\textbf{Fine-Tuning}} & \multirow{2}{*}{\textbf{\spot{}}} & \multirow{2}{*}{\textbf{\pth{}}} & \multirow{2}{*}{\textbf{\mixture}} \\ \cmidrule(lr){2-5}
 & \tfive{} & \ktfive{} & \tfive{} & \ktfive{} &  &  &  \\ \midrule
\textsc{large} & 4.47 & 5.29 & 10.03 & \textbf{11.19} & 7.22 & 8.62 & 6.58 \\
\textsc{xxl} & 6.42 & 8.62 & 12.92 & 13.47 & 20.03 & \textbf{29.37} & 23.09 \\ \bottomrule
\end{tabular}%
\caption{Test set EM scores achieved by various tuning methods on 2WikiMultiHopQA \citep{ho-etal-2020-constructing}. \spot{} \citep{vu-etal-2022-spot}, \pth{}, and \mixture{} use \ktfive{} as their base model.}
\label{tab:twohopresults}
\vspace{-1em}
\end{table*}
To this end, we compute the EM and F1 scores over the full generated spans of entities, interleaved by the relations that connect them. 
Note that EM is substantially stricter than F1, since F1 rewards partial overlap of tokens between the target vs. the generated output.
\Cref{tab:hpmetrics} shows these scores for \ktl{} and \ktxxl{} on the validation set of our random walk task, tuned using the \hp{}. 
We see from \Cref{tab:hpmetrics} that there is a substantial gap between \ktl{} ($\sim$23 EM) and \ktxxl{} ($\sim$58 EM), suggesting that the \textsc{large} model finds it difficult to generalize to random walk paths involving entities and relations outside of the training set.
We conclude from this observation that the gap between \ktl{} and \ktxxl{} in generalizing to held-out KG paths is likely going to be reflected when tested for 2-hop QA. That is, we expect our prompting methods with \ktl{} as the base-model to struggle on our test set questions as their ground-truth paths were not encountered during training, and at the same time, expect the opposite to be the case for \ktxxl{}.
Additionally, the EM score achieved by the \textsc{XXL}-sized model is well below perfect values, highlighting important avenues for future work to improve upon these gaps. 

\paragraph{Training on random walks substantially improves 2-hop capabilities ..but mostly in larger LMs}

We used three methods that leveraged the training signal provided by random walks to compose the 1-hop knowledge as memorized by \ktfive{}: \pth{} (ours), \mixture{} (ours), and \spot{} \citep{vu-etal-2022-spot}.
Due to lack of space, examples of the outputs from each of these methods, along with analysis of intermediate steps (e.g., parsing) are shown in Appendix \ref{sec:aa}.
We observe from \Cref{tab:twohopresults} that for the \textsc{xxl}-sized model, all three methods lead to substantial improvements in performance on 2-hop questions over standard tuning approaches on T5 and \ktfive{}.
Notably for \ktxxl{}, random walk-integrated methods improve even over fine-tuning, which is often expected to be better at transfer learning as compared to parameter efficient methods. 
Among the three, our \pth{} method shows the best improvements ($\sim$16 point gain over fine-tuning \ktxxl{}) at answering 2-hop questions. 
This showcases the promise of learning separate specialized prompts that operate over the same underlying model to first parse natural language into incomplete structured knowledge, and then expand it to answer the question, while also eliciting intermediate steps \citep{wang2022iteratively}, similar to recent in-context prompting methods \citep{wei2022chain, nye2022show}.
While the \mixture{} method ($\sim$9.6 point gain over fine-tuning) falls short of \pth{}, it still improves over \spot{} ($\sim$6.6 point gain over fine-tuning), suggesting that joint training of related tasks may improve over sequential training (as employed by \spot) in performing multi-hop reasoning, at larger model sizes.
In the case of \tl{} and \ktl{}, while the proposed methods show improvements over standard prompt-tuning, with \pth{} demonstrating a gain of 3.33 points over prompt-tuning \ktl{}, they fall-short of the performance achieved by fine-tuning. 
However, their non-trivial improvements over regular prompt-tuning suggests the general benefits of the training signal provided by random walks, which end up being most impressive at models that are an order of magnitude larger. Overall, these results corroborate with our hypothesis from the random walk tests about \ktl{}'s potential inability to generate partially novel random walks given either natural language multi-hop questions (\mixture) or their parses (\pth).

\section{Conclusion}
We show that composition of memorized world knowledge can be triggered in LMs with up to 11B parameters (\txxl{}) to a desirable extent by leveraging training signal from random walks over structured knowledge using approaches based on prompt-tuning \citep{lester-etal-2021-power}. Doing so leads to substantial improvements in the LMs' ability to answer 2-hop questions, even beyond standard, full model fine-tuning.

\section*{Limitations}
Despite showing non-trivial improvements in the multi-hop capabilities of T5 models, our work has multiple limitations.

\paragraph{Restricted to 2-hops} First, we chose 2WikiHopMultiQA \citep{ho-etal-2020-constructing} as our primary dataset since it uniquely maps each question to a chain of triples that contain the precise, noiseless single-hop knowledge required to answer the question. However, this comes at the cost of our analyses only being restricted to 2-hops (though see arguments by \citet[sec 3.5]{press2022measuring} who suggest 3-and-4-hop questions to be too convoluted to understand even by native-speakers). 
Nonetheless, our random walk training method is general by definition, and can be extended to multiple hops, though its effectiveness on QA tasks requiring more than 2-hops of reasoning remains to be measured. 

\paragraph{Knowledge Graph size}
Our focus in this paper was to allow models to chain together their internalized knowledge in order to answer complex 2-hop questions. However, this critically requires them to possess the world knowledge required to answer the questions, for which we had to memorize the KG constructed using the structured triples provided in the dataset. 
This trade-off between focusing on knowledge composition vs. fully encoding world knowledge restricted our KG to be small in size (only \ents{} entities and 29 relations), which could be impractical in most real-world applications. 
In future work, we will experiment with larger sized KGs \citep{vrandevcic2014wikidata}, by adding a substantially larger amount of additional triples to the existing KG, and measure their impact on multi-hop reasoning.

\paragraph{Lack of diverse QA tasks} Finally, we were unable to consider popular datasets with CBQA versions such as TriviaQA \citep{roberts-etal-2020-much}, NaturalQuestions \citep{kwiatkowski-etal-2019-natural}, etc., due to their lack of links from questions to structured knowledge. 
Future work can apply entity and relational linking techniques \citep{balachandran-etal-2021-investigating, agarwal-etal-2021-knowledge} in order to augment such QA datasets with (possibly) noisy links to structured knowledge, which will allow us to paint a more holistic picture of our methods. Additionally, this would also overcome the above limitation (of KG size), as it would substantially increase the amounts of entities and relations to be encoded within models.

\paragraph{Implications for Larger Models} 
Although we show clear improvements in triggering 2-hop reasoning in the largest T5 LM (\txxl{}), with 11B parameters, contemporary work has shown that multi-step reasoning capacities naturally emerge in LMs that are two or three orders of magnitude larger \citep{brown2020language, chowdhery2022palm, wei2022chain, wei2022emergent}. However, these LMs benefit from examples in-context (especially since tuning them is non-trivial and expensive), and therefore it is unclear whether our methods can improve such models’ capacities even further. We have not tested such LMs in our work, due to resource limitations.

\section*{Acknowledgments}
We thank Noah Constant, Chung-Ching Chang, Brian Lester, and Ben Withbroe from Google Research for their helpful comments and advice. We would also like to thank our three anonymous reviewers for their useful feedback.

\bibliography{anthology,custom}

\appendix

\section{Training and Experiment Details}
\label{sec:trainingdeets}
\paragraph{Hyperparameters} We use the default hyperparameters and optimizers used to train the T5 1.1 checkpoints \citep{raffel2020exploring} as well as those used in the Prompt-Tuning and \spot{} papers \citep{lester-etal-2021-power, vu-etal-2022-spot}. We set the prompt-length to 100 for all prompt-tuning experiments, and initialized them with the top 100 tokens in the T5 models' vocabulary, following \citet{lester-etal-2021-power}. We fine-tune and prompt-tune our models for a maximum of 100K and 200K steps, respectively. We stop training on convergence, and use the checkpoint with the best validation performance to evaluate. 
Tables \ref{tab:finetuning}, \ref{tab:knowledgeintegration}, and  \ref{tab:prompttuning} show hyperparameter values for each type of experiment. All results are from single runs.

\paragraph{Hardware and Compute} Prompt-tuning and fine-tuning experiments for \textsc{large} models were run on 16 TPUv3 chips, while those for \textsc{xxl} models were run on 64 TPUv3 chips. One exception is knowledge integration (which also involved continual pre-training on C4, larger batch size, and longer sequences), for which we used 256 TPUv3 chips for \textsc{xxl}, and 64 TPUv3 chips for \textsc{large}.

\paragraph{Code} For metric calculation and checkpoints, we use the T5 and T5x code-base, open-sourced on github.\footnote{\url{https://github.com/google-research/text-to-text-transfer-transformer/tree/main/t5}}\footnote{\url{https://github.com/google-research/t5x}} For prompt-tuning experiments, we adapt the original code-base \citep{lester-etal-2021-power}, which is also open-sourced.\footnote{\url{https://github.com/google-research/prompt-tuning}}

\paragraph{Data} The 2WikiMultiHopQA dataset \citep{ho-etal-2020-constructing} has been released with Apache 2.0 license.\footnote{\url{https://github.com/Alab-NII/2wikimultihop}}

\begin{table}[ht]
\centering
\resizebox{0.8\columnwidth}{!}{%
\begin{tabular}{@{}lr@{}}
\toprule
\textbf{Hyperparameter} & \textbf{Values} \\ \midrule
Batch Size & 32 (\textsc{xxl}), 128 (\textsc{large}) \\
Learning Rate & 0.001 \\
Dropout & 0.1 \\
Training Steps & 100K (w/ early stopping) \\ \bottomrule
\end{tabular}%
}
\caption{Hyperparameters used for fine-tuning \tl{} and \txxl{}. Values except batch size and training steps kept same as \citet{raffel2020exploring}.}
\label{tab:finetuning}
\end{table}
\begin{table}[h]
\centering
\resizebox{0.8\columnwidth}{!}{%
\begin{tabular}{@{}lr@{}}
\toprule
\textbf{Hyperparameter} & \textbf{Values} \\ \midrule
Batch Size & 512 \\
Learning Rate & 0.001 \\
Dropout & 0.1 \\
Training Steps & 100K (w/ early stopping) \\ \bottomrule
\end{tabular}%
}
\caption{Hyperparameters used for Knowledge Integration experiments. Values except batch size and training steps kept same as \citet{raffel2020exploring}.}
\label{tab:knowledgeintegration}
\end{table}
\begin{table}[h]
\centering
\resizebox{0.8\columnwidth}{!}{%
\begin{tabular}{@{}lr@{}}
\toprule
\textbf{Hyperparameter} & \textbf{Values} \\ \midrule
Batch Size & 32 (\textsc{xxl}), 128 (\textsc{large}) \\
Learning Rate & 0.3 \\
Prompt Length & 100 \\
Dropout & 0.1 \\
Training Steps & 200K (w/ early stopping) \\ \bottomrule
\end{tabular}%
}
\caption{Hyperparameters used for all prompt-tuning experiments. Values except batch size kept same as \citet{lester-etal-2021-power}, number of training steps kept same as \citet{vu-etal-2022-spot}, who found longer training to be beneficial.}
\label{tab:prompttuning}
\end{table}

\section{Additional Analyses}
\label{sec:aa}

\subsection{Knowledge Integration}
\label{sec:aaki}
\begin{figure}[t!]
    \centering
    \includegraphics[width=0.8\columnwidth]{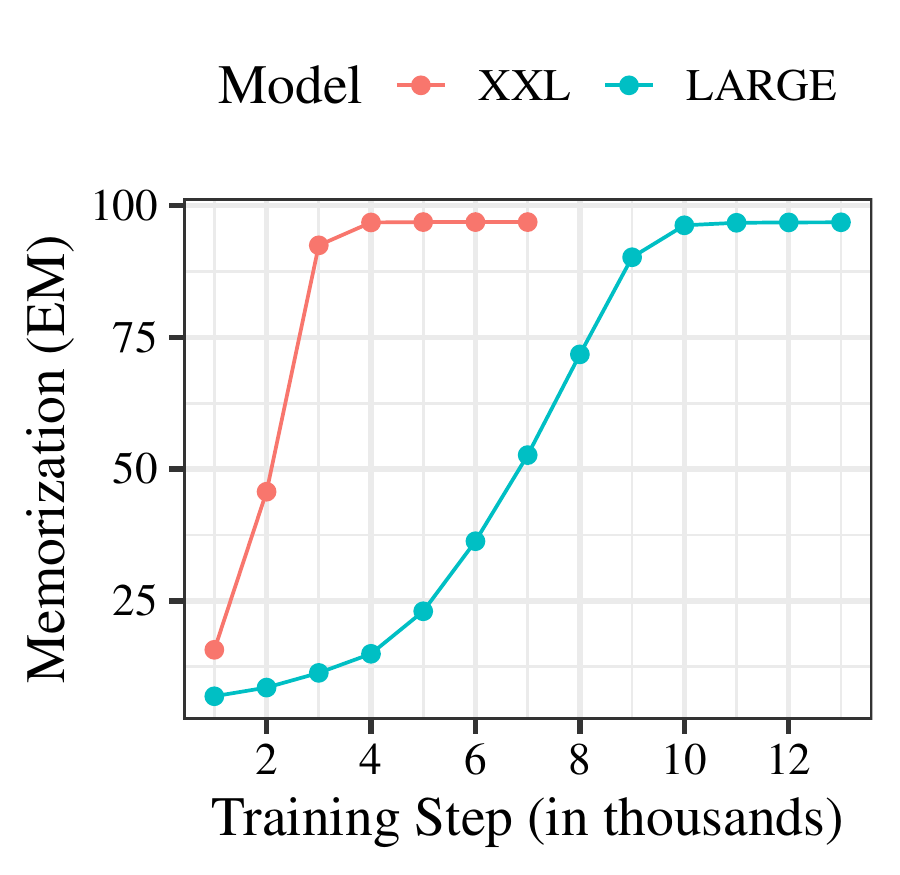}
    \caption{Time course of KG memorization for different \ktfive{} model sizes. EM scores calculated for producing object entity ($e_2$), given subject ($e_1$) and relation ($r$) as inputs to T5 models.}
    \label{fig:kgmem}
\end{figure}

Integrating single-hop entity knowledge is an important part of our methods. How well are the models able to actually encode this knowledge? \Cref{fig:kgmem} shows the dynamics of memorization across both models, measured as the exact match scores in generating $e_2$ given $e_1$ and $r$. From \Cref{fig:kgmem}, we see that the \textsc{xxl} and \textsc{large} models can memorize 96\% of the KG within 5,000 and 10,000 steps respectively. With a batch size of 512, this translates to traversing the dataset 27 and 54 times, respectively, for \textsc{xxl} and \textsc{large}. An important caveat here is that the models are also being tuned on C4 \cite{raffel2020exploring}, in order to retain the models' general language understanding-like capabilities. 
That is, they can be expected to memorize the KG relatively faster in the absence of training on the C4 corpus, but this would constitute a trade-off, by leading to overfitted models with substantial loss their original utility on other NLP tasks.

\subsection{Parsing Step in \pth{}}
The parsing step is essential for our \modular{} approach to succeed. Here we perform additional analyses on how well models can successfully extract the relational structure that is required to answer the 2-hop questions in 2WikiMultiHopQA. 
Recall that the objective of the parsing step is to produce as output a sequence indicating an incomplete random walk, containing only the initial entity (seed node), followed by the relations (edges) that lead to the final entity. For instance, if the question is ``\textit{Where was the director of Inception (film) born?}'' the output of the parsing step should be:
\begin{quote}
    \texttt{Inception (film) ;  director ; place of birth}
\end{quote}
Here, \texttt{Inception (film)} is the entity, $e_1$, while \texttt{director} and \texttt{place of birth} are the relations, $r_1$ and $r_2$, respectively. 
We analyze the extent to which models successfully extract these three elements for the \ntest{} test set questions, by measuring three quantities: (1) \textbf{Relation EM}, which is the exact match score computed between the ground truth span of relation pairs (here ``\texttt{director ; place of birth}''), and that extracted from the model outputs; (2) \textbf{Entity EM}, which is similar to Relation EM, but only considers the initial entity; and (3) \textbf{Full EM}, which computes the exact match score between the full output and the target. \Cref{tab:parsingmetrics} shows these values from prompt-tuning the two \ktfive{} models.
\begin{table}[t!]
\centering
\resizebox{\columnwidth}{!}{%
\begin{tabular}{@{}llll@{}}
\toprule
\textbf{Model} & \textbf{Relation EM} & \textbf{Entity EM} & \textbf{Full EM} \\ \midrule
\ktl{} & 98.69 & 76.19 & 78.98 \\
\ktxxl{} & 99.17 & 78.46 & 80.17 \\ \bottomrule
\end{tabular}%
}
\caption{Metrics for the parsing sub-task of \pth{} on test-set questions.}
\label{tab:parsingmetrics}
\end{table}

From \Cref{tab:parsingmetrics}, we see that prompt-tuning both models allows them to achieve almost perfect EM values in extracting the relation pairs from the questions. However, we notice that models are not able to maintain this performance in copying over the entity, which lowers their overall EM scores on this task. 
We performed a manual analysis of 50 randomly sampled outputs---with incorrect entity predictions---and found most errors to be due to omission of tokens involving middle names, or additional information about the entity such as the ``\texttt{(film)}'' in the above example (other examples include the entity's title, such as ``\texttt{Count of East Frisia}'', or \texttt{``(born in year XXX)''}, \texttt{``(died in year XXX)''}, etc.)

\subsection{Example Outputs}
Tables \ref{tab:example1}, \ref{tab:example2}, \ref{tab:example3}, and \ref{tab:example4} show examples of outputs from the different approaches used in this work (examples shown for the \textsc{xxl}-sized models). Below we discuss each of these cases in detail:

\begin{itemize}
    \item In \Cref{tab:example1}, all approaches that leverage the training signal from random walks succeed, while tuning methods that do not fail. Additionally, all three random walk-integrated methods agree on their parsed relational structure as well as the intermediate entity.
    \item In \Cref{tab:example2}, only the two proposed methods (\pth{} and \mixture{}) succeed, while all other methods fail. Note that \spot{} correctly predicts the correct intermediate entity (\texttt{Sally Hemings}), but is unable to predict the final entity (\texttt{John Wayles}).
    \item \Cref{tab:example3} shows an example where all approaches fail. However, this question is ambiguous, as \textit{aunt} can either mean \textit{father's sister} or \textit{mother's sister} -- our random walk integrated methods correctly predict these relational structures but are unable to resolve the intermediate and final entities.
    \item \Cref{tab:example4} shows an example where all approaches are supposedly scored as incorrect, but are in-fact correct. Here we argue that the ground truth answer, ``\textit{United Kingdom}'' is in its incorrect form, since the question asks for the nationality of a person. Our random walk-integrated methods successfully predict the relational structure and intermediate entities. Moreover all approaches predict \texttt{British} or \texttt{English}, which are more acceptable forms of nationality for persons from the United Kingdom. This problem could be mitigated by adding in aliases for the entities in the ground-truth answer space, similar to TriviaQA \citep{roberts-etal-2020-much}.
\end{itemize}

\renewcommand*{\arraystretch}{1.2}
\begin{table*}[ht!]
\centering
\resizebox{\linewidth}{!}{%
\begin{tabular}{@{}lll@{}}
\toprule
\multicolumn{3}{l}{\textbf{Question:} \textit{Where was the place of burial of the director of film New World (1995 Film)?} \textbf{Answer:} Père Lachaise Cemetery} \\ \midrule
\textbf{Model} & \textbf{Setup} & \textbf{Response} \\ \midrule
\multirow{2}{*}{\txxl{}} & FT & \texttt{\wrongans{Forest Lawn Memorial Park}} \\
 & PT & \texttt{\wrongans{Forest Lawn Memorial Park}} \\ \midrule
\multirow{6}{*}{\ktxxl{}} & FT & \texttt{\wrongans{New York}} \\
 & PT & \texttt{\wrongans{Forest Lawn Memorial Park}} \\
 & \spot & \texttt{New World ; director ; Alain Corneau ; place of burial ; \correctans{Père Lachaise Cemetery}} \\
 & \multirow{2}{*}{\pth} & \textbf{PP:} \texttt{New World ; director ; place of burial} \\
 &  & \textbf{HP:} \texttt{New World ; director ; Alain Corneau ; place of burial ; \correctans{Père Lachaise Cemetery}} \\
 & \mixture & \texttt{New World ; director ; Alain Corneau ; place of burial ; \correctans{Père Lachaise Cemetery}} \\ \bottomrule
\end{tabular}%
}
\caption{An example case where methods that leverage random walks succeed, but baselines fail.}
\label{tab:example1}
\end{table*}
\renewcommand*{\arraystretch}{1.2}
\begin{table*}[ht!]
\centering
\resizebox{\textwidth}{!}{%
\begin{tabular}{@{}lll@{}}
\toprule
\multicolumn{3}{l}{\textbf{Question:} \textit{Who is Harriet Hemings's maternal grandfather?} \textbf{Answer:} John Wayles} \\ \midrule
\textbf{Model} & \textbf{Setup} & \textbf{Response} \\ \midrule
\multirow{2}{*}{\txxl{}} & FT & \texttt{\wrongans{Ted Hughes}} \\
 & PT & \texttt{\wrongans{John Hemings}} \\ \midrule
\multirow{6}{*}{\ktxxl{}} & FT & \texttt{\wrongans{Betty Hemings}} \\
 & PT & \texttt{\wrongans{John Hemings}} \\
 & \spot & \texttt{Harriet Hemings ; mother ; Sally Hemings ; father ; \wrongans{Thomas Hemings}} \\
 & \multirow{2}{*}{\pth} & \textbf{PP:} \texttt{Harriet Hemings ; mother ; father} \\
 &  & \textbf{HP:} \texttt{Harriet Hemings ; mother ; Sally Hemings ; father ; \correctans{John Wayles}} \\
 & \mixture & \texttt{Harriet Hemings ; mother ; Sally Hemings ; father ; \correctans{John Wayles}} \\ \bottomrule
\end{tabular}%
}
\caption{An example case where all baselines fail, and additionally SPoT \citep{vu-etal-2022-spot} also produces the incorrect final entity, but our two proposed methods succeed.}
\label{tab:example2}
\end{table*}
\renewcommand*{\arraystretch}{1.2}
\begin{table*}[ht!]
\centering
\resizebox{\textwidth}{!}{%
\begin{tabular}{@{}lll@{}}
\toprule
\multicolumn{3}{l}{\textbf{Question:} \textit{Who is Christopher Blom Paus's aunt?} \textbf{Answer:} Hedevig Christine Paus} \\ \midrule
\textbf{Model} & \textbf{Setup} & \textbf{Response} \\ \midrule
\multirow{2}{*}{\txxl{}} & FT & \texttt{\wrongans{Clotilde of Saxe - Lauenburg}} \\
 & PT & \texttt{\wrongans{Annemarie Blom Paus}} \\ \midrule
\multirow{6}{*}{\ktxxl{}} & FT & \texttt{\wrongans{Anna of Oldenburg}} \\
 & PT & \wrongans{\texttt{Christina Paus}} \\
 & \spot & \texttt{Christopher Blom Paus ; father ; Ole Paus ; sibling ; \wrongans{Kjersti Bua Paus}} \\
 & \multirow{2}{*}{\pth} & \textbf{PP:} \texttt{Christopher Blom Paus ; mother ; sibling} \\
 &  & \textbf{HP:} \texttt{Christopher Blom Paus ; mother ; Margrete Laarmann ; sibling ; \wrongans{Kjartan Flóki}} \\
 & \mixture & \texttt{Christopher Blom Paus ; mother ; Ulla Blom ; sibling ; \wrongans{Gunnar Blom}} \\ \bottomrule
\end{tabular}%
}
\caption{An example of an ambiguous question (since ``aunt'' can be father's sister or mother's sister) on which all approaches fail. Importantly, methods that use random-walks accurately generate the relations required to answer the question, but fail at predicting the correct entities.}
\label{tab:example3}
\end{table*}
\renewcommand*{\arraystretch}{1.2}
\begin{table*}[ht!]
\centering
\resizebox{\textwidth}{!}{%
\begin{tabular}{@{}lll@{}}
\toprule
\multicolumn{3}{l}{\textbf{Question:} \textit{What nationality is John Bede Dalley's father ?} \textbf{Answer:} United Kingdom} \\ \midrule
\textbf{Model} & \textbf{Setup} & \textbf{Response} \\ \midrule
\multirow{2}{*}{\txxl{}} & FT & \texttt{\textcolor{OliveGreen}{\textbf{British}}} \\
 & PT & \texttt{\textcolor{OliveGreen}{\textbf{British}}} \\ \midrule
\multirow{6}{*}{\ktxxl{}} & FT & \texttt{\textcolor{OliveGreen}{\textbf{English}}} \\
 & PT & \texttt{\textcolor{OliveGreen}{\textbf{English}}} \\
 & \spot & \texttt{John Bede Dalley ; father ; William Dalley ; country of citizenship ; \textcolor{OliveGreen}{\textbf{English}}} \\
 & \multirow{2}{*}{\pth} & \textbf{PP:} \texttt{John Bede Dalley ; father ; country of citizenship} \\
 &  & \textbf{HP:} \texttt{John Bede Dalley ; father ; William Bede Dalley ; country of citizenship ; \textcolor{OliveGreen}{\textbf{English}}} \\
 & \mixture & \texttt{John Bede Dalley ; father ; William Dalley, 1st Viscount Darnley ; country of citizenship ; \textcolor{OliveGreen}{\textbf{British}}} \\ \bottomrule
\end{tabular}%
}
\caption{An example of a scenario where all models fail at answering the question correctly, but this is likely attributable to the dataset since it does not contain aliases.}
\label{tab:example4}
\end{table*}

\begin{table*}[!t]
\centering
\resizebox{\linewidth}{!}{%
\begin{tabular}{@{}p{0.15\textwidth}p{0.35\textwidth}p{0.15\textwidth}p{0.35\textwidth}@{}}
\toprule
\textbf{Relation} & \textbf{Template Space} & \textbf{Relation} & \textbf{Template Space} \\ \midrule
\texttt{director} & \textit{Who is the director of X?, Who directed the film X?} & \texttt{mother} & \textit{Who is the mother of X?, Who is X's mother?} \\
\texttt{date of birth} & \textit{What is the date of birth of X?, When is X's birthday?, When was X born?} & \texttt{founded by} & \textit{Who is the founder of X?, Who founded X?} \\
\texttt{date of death} & \textit{When did X die?, What is the date of death of X?} & \texttt{inception} & \textit{When was X founded?} \\
\texttt{country} & \textit{What country is X from?, What is the nationality of X?} & \texttt{manufacturer} & \textit{Who manufactures X?} \\
\texttt{country of citizenship} & \textit{What country is X from?, What is the nationality of X?} & \texttt{performer} & \textit{Who is the performer of the song X?, Who performed the song X?} \\
\texttt{award received} & \textit{What is the award that X received?, Which award did X receive?} & \texttt{place of birth} & \textit{Where was X born?, What is the place of birth of X?} \\
\texttt{cause of death} & \textit{Why did X die?, What was the cause of X's death?} & \texttt{place of burial} & \textit{Where was X buried?, Where is the place of burial of X?} \\
\texttt{composer} & \textit{Who is the composer of X?, Who composed X?} & \texttt{place of death} & \textit{Where did X die?, Where is the place of death of X?} \\
\texttt{creator} & \textit{Who is the creator of X?, Who created X?} & \texttt{place of detention} & \textit{Where did X go to prison?, Where was X detained?} \\
\texttt{child} & \textit{Who is the child of X?} & \texttt{presenter} & \textit{Who is the presenter of X?, Who presented X?} \\
\texttt{doctoral advisor} & \textit{Who is the doctoral advisor of X?} & \texttt{publisher} & \textit{Who published X?, What company published X?} \\
\texttt{editor} & \textit{Who is the editor of X?, Who edited X?} & \texttt{sibling} & \textit{Who is the sibling of X?, Who is X's sibling?} \\
\texttt{educated at} & \textit{Where did X graduate from?, What is the alma mater of X?, Where did X study?} & \texttt{spouse} & \textit{Who is the spouse of X?, Who is X's spouse?} \\
\texttt{employer} & \textit{Who is the employer of X?, Where does X work?} & \texttt{student of} & \textit{Who was the teacher of X?, Who was X's teacher?} \\
\texttt{father} & \textit{Who is the father of X?, Who is X's father?} &  & \textit{} \\ \bottomrule
\end{tabular}%
}
\caption{Question templates for for each of the 29 relations, used to create \onehopqa{}. \textit{X} stands for the subject.}
\label{tab:onehopq}
\end{table*}

\section{Templates for constructing \onehopqa{}}
\label{sec:onehop}

Here we describe our process of constructing \onehopqa{}: a collection of English question-answer pairs that only require single-hop knowledge using the 2WikiMultiHopQA \citep{ho-etal-2020-constructing} dataset.
The 2WikiMultiHopQA dataset provides unique sequences of single-hop triples that collectively answer each 2-hop question. These amount to a total of 95,103 unique triples spanning \ents{} unique entities and 29 relations. We manually define a diverse set of templates for each relation, as shown in Table \ref{tab:onehopq}. For many relations, we have multiple different paraphrases of the question template, e.g., the relation \texttt{director} translates to: \textit{Who is the director of X?} or \textit{Who directed the film X?} In such cases, we randomly sample a template from the entire set, equally weighing each. In total, we end up with 83,643 train, 5,022 validation, and 6,440 test QA pairs.

\end{document}